\documentclass[11pt,a4paper]{article}
\usepackage[hyperref]{style/acl2021}
\usepackage{times}
\usepackage{latexsym}
\usepackage{diagbox}

\usepackage{microtype}

\aclfinalcopy % Uncomment this line for the final submission
\def\aclpaperid{112} %  Enter the acl Paper ID here

\newif\ifcomment\commenttrue
\usepackage{framed}
\usepackage{mdwlist}
\usepackage{latexsym}
\usepackage{colortbl}
\usepackage{xcolor}
\usepackage{nicefrac}
\usepackage{booktabs}
\usepackage{amsfonts}
\usepackage[T1]{fontenc}
\usepackage{bold-extra}
\usepackage{amsmath}
\usepackage{amssymb}
\usepackage{bm}
\usepackage{graphicx}
\usepackage{mathtools}
\usepackage{microtype}
\usepackage{multirow}
\usepackage{multicol}
\usepackage{xspace}
\usepackage{latexsym,comment}
\usepackage{pifont}% http://ctan.org/pkg/pifont
\usepackage{listings}

\newcommand{\gem}[1]{\mbox{\textsc{gem}}}
\newcommand{\abr}[1]{\textsc{#1}}

\newcommand{\hidetext}[1]{}
\newcommand{\ignore}[1]{}

\ifcomment
\newcommand{\pinaforecomment}[3]{\colorbox{#1}{\parbox{.8\linewidth}{#2: #3}}}
\else
\newcommand{\pinaforecomment}[3]{}
\fi

\newcommand{\smallurl}[1]{ \begin{tiny}\url{#1}\end{tiny}}

\definecolor{lightblue}{HTML}{3cc7ea}
\definecolor{CUgold}{HTML}{CFB87C}
\definecolor{grey}{rgb}{0.95,0.95,0.95}
\definecolor{ceil}{rgb}{0.57, 0.63, 0.81}
\definecolor{UMDred}{HTML}{ed1c24}
\definecolor{UMDyellow}{HTML}{ffc20e}

\newif\ifsubscripterror\subscripterrortrue
\ifsubscripterror
\newcommand{\err}[1]{\textsubscript{~$\pm$#1}}
\else
\newcommand{\err}[1]{ $\pm$ #1}
\fi

\newcommand\masklm{\textsc{MaskLM}\xspace}

\newcommand\sqa{\textsc{SQA}\xspace}
\newcommand\tabfact{\textsc{TabFact}\xspace}

\newcommand{\tapas}{\textsc{Tapas}\xspace}

\newcommand{\bert}{\textsc{Bert}\xspace}

\newcommand{\semtabfact}{\textsc{SemTabFact}\xspace}
\newcommand{\infotabs}{\textsc{InfoTabs}\xspace}

\title{TAPAS at SemEval-2021 Task 9: Reasoning over tables with intermediate pre-training}

\author{Thomas M{\"u}ller, Julian Martin Eisenschlos, Syrine Krichene \\ \\
  Google Research, Z{\"u}rich \\
  \texttt{\{thomasmueller,eisenjulian,syrinekrichene\}@google.com}}

\date{}

\begin{document}

\maketitle

\begin{abstract}

We present the TAPAS contribution to the Shared Task on Statement Verification and Evidence Finding with Tables (SemEval 2021 Task 9, \citet{wang-etal-2021-semeval}). \semtabfact{} Task A is a classification task of recognising if a statement is entailed, neutral or refuted by the content of a given table.
We adopt the binary \tapas{} model of \citet{eisenschlos-etal-2020-understanding} to this task. We learn two binary classification models: A first model to predict if a statement is neutral or non-neutral and a second one to predict if it is entailed or refuted. As the shared task training set contains only entailed or refuted examples, we generate artificial neutral examples to train the first model. Both models are pre-trained using a \masklm objective, intermediate counter-factual and synthetic data \cite{eisenschlos-etal-2020-understanding} and \tabfact~\cite{Chen2020TabFact}, a large table entailment dataset.
We find that the artificial neutral examples are somewhat effective at training the first model, achieving 68.03 test F1 versus the 60.47 of a majority baseline.
For the second stage, we find that the pre-training on the intermediate data and \tabfact{} improves the results over \masklm pre-training (68.03 vs 57.01).
\end{abstract}

\section{Introduction}
\label{sec:intro}

% What is the task about and why is it important? Be sure to mention the language(s) covered and cite the task overview paper. ~1 paragraph 
Recently, the task of Textual Entailment (TE) \cite{dagan2005pascal} or Natural Language Inference (NLI) \cite{bowman-etal-2015-large}
has been adapted to a setup where the premise is a table \cite{Chen2020TabFact,gupta2020infotabs}.
The Shared Task on Statement Verification and Evidence Finding with Tables (SemEval 2021 Task 9, \citet{wang-etal-2021-semeval}) follows this line of work and provides a new dataset
consisting of tables extracted from scientific articles and natural language statements written by crowd workers.
In this paper, we discuss a system for tackling task A, which is a multi-class classification task that requires finding if a statement is entailed, neutral or refuted by the contents of a table.
The training set contains only entailed and refuted examples and requires data augmentation to learn the neutral class.
Additionally, this data set is composed of English language data and requires sophisticated contextual and numerical reasoning such as handling comparisons and aggregations.

\begin{figure}[t]
    \centering
    \includegraphics[width=.9\linewidth]{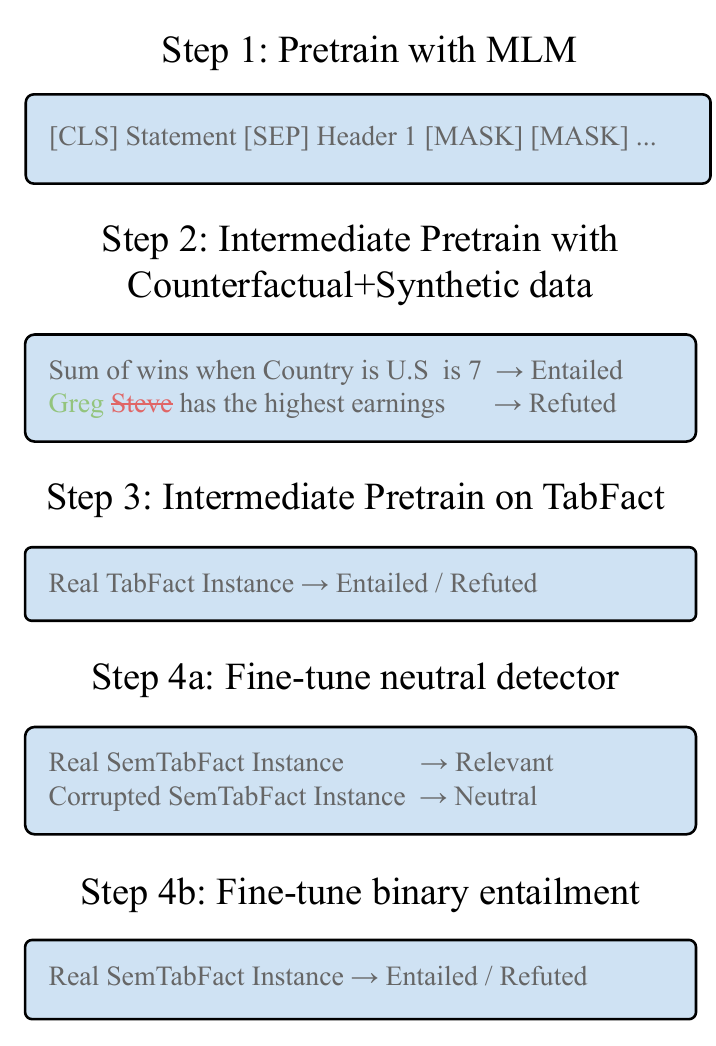}
    \caption{Overview of the training pipeline use in our system. We use intermediate pre-training on \emph{Counterfactual+Synthetic} data~\cite{eisenschlos-etal-2020-understanding} and then fine-tune on \tabfact~\cite{Chen2020TabFact}.}
    \label{fig:pipeline}
\end{figure}

A successful line of research on table entailment \cite{Chen2020TabFact,eisenschlos-etal-2020-understanding,gupta2020infotabs} has been driven by \bert-based models \cite{devlin-etal-2019-bert}. 
These approaches reason over tables without generating logical forms to directly predict the entailment decision.
Such models are known to be efficient on representing textual data as well reasoning over semi-structured data such as tables.
In particular, \tapas-based models \cite{herzig-etal-2020-tapas} that encode the table structure using additional embeddings, have been successfully used to solve binary entailment tasks with tables \cite{eisenschlos-etal-2020-understanding}.

% What is the main strategy your system uses? ~1 paragraph pretraining + 2 tage model.

\begin{figure}[ht]
    \centering
    \includegraphics[width=.75\linewidth]{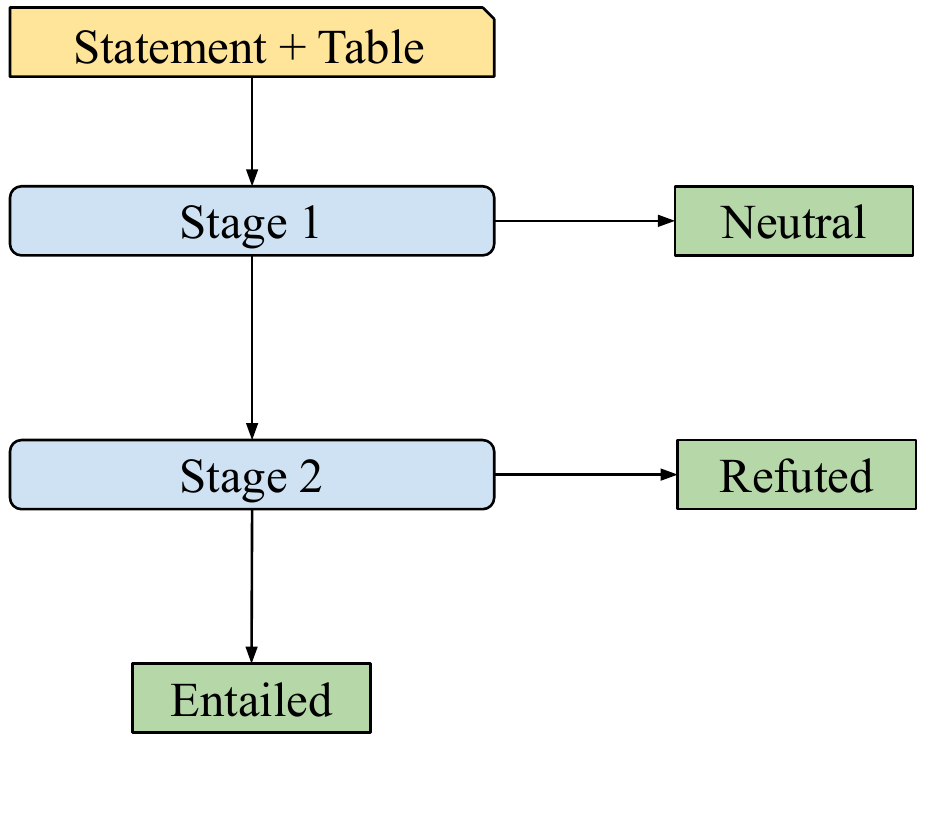}
    \caption{Overview of the complete system. Stage 1 classifies into neutral and non-neutral statements. Stage 2 into entailed and refuted. Both stages are based on binary \tapas classifier models.}
    \label{fig:system}
\end{figure}

To address multi-class classification entailment we decompose the main task into two sub-tasks and use two \tapas models as described in Figure~\ref{fig:system}. A first model classifies the statement into neutral or non-neutral, and a second into entailed or refuted.
The two models are learned separately: we created artificial neutral statements to fine-tune the first model.
Examples are extracted by randomly pairing statements and tables from the \semtabfact{} training set.
We also generate harder examples by creating new tables from the original tables by removing columns that contain evidence to refute or entail the statement.
This procedure is discussed in Section \ref{sec:system}.

We follow \citet{eisenschlos-etal-2020-understanding} and pre-train the two \tapas models with a \masklm objective \cite{devlin-etal-2019-bert} and then with counterfactual and synthetic data as shown in Figure~\ref{fig:pipeline}. We additionally fine-tune both models on the \tabfact dataset. Details are given in Section \ref{sec:experiments}.

We find that our artificial neutral statement creations out-performs a majority baseline and that pre-training help for both the first and the second stage.
Our best models achieve 68.03 average micro f1-score on the test set. 

\section{Related Work}
\label{sec:related}

\paragraph{Entailment on Tables}
Recognizing textual entailment~\cite{dagan-10} has expanded from a text only task to incorporate more structured data, such knowledge graphs \cite{vlachos-riedel-2015-identification}, tables \cite{jo2019aggchecker,gupta2020infotabs} and images \cite{suhr2017corpus,suhr2019corpus}.

The \tabfact~dataset~\cite{Chen2020TabFact} for example, uses tables as the \emph{premise}, or source of information to resolve whether a statement is entailed or refuted. 
The \tapas architecture introduced by \citet{herzig-etal-2020-tapas} can be used to obtain transformer-based baselines, as shown in ~\citet{eisenschlos-etal-2020-understanding}, by using special embeddings to encode the table structure. ~\citet{zhang-etal-2020-table, Chen2020TabFact} also use \bert like models but obtain less accurate results due possibly to not using table-specific pre-training.

\paragraph{Intermediate Pre-training}
Our system relies on intermediate pre-training, a technique that appears in different forms in the literature. Language model fine-tuning \cite{howard-2018-universal}, or domain adaptive pre-training \cite{gururangan-2020-dont-stop} are useful applications for domain adaptation.
In a similar manner than \citet{pruksachatkun-2020-intermediate-task}, we use the \emph{Counterfactual+Synthetic} tasks from ~\citet{eisenschlos-etal-2020-understanding} to improve the discrete and numeric reasoning capabilities of the model for Table entailment.

\paragraph{Synthetic data} The use of synthetic data to improve learning in \abr{nlp} is ubiquitous \cite{alberti-etal-2019-synthetic,lewis-etal-2019-unsupervised,wu-etal-2016-bilingually, leonandya-etal-2019-fast}.
\citet{salvatore-etal-2019-logical} focus on textual entailment and probes models with synthetic examples.
In semantic parsing \citet{wang-etal-2015-building, iyer-17, dbpal} use templates to augment the training data for text-to-SQL tasks and \citet{Geva2020InjectingNR} do so to improve numerical reasoning, as do~\citet{eisenschlos-etal-2020-understanding} on tabular data. They also create minimal contrastive examples~\cite{Kaushik2020Learning,gardner-2020} by automatically swapping entities in the statements by plausible alternatives that exists elsewhere in the table.

\begin{table*}
    \centering
    \begin{tabular}{l|rrrrr}
    \toprule
         Dataset & Statements & Tables & Entailed & Refuted & Neutral \\
         \midrule
         Crowdsourced Train & 4,506 & 981 & 2,818 (62.54\%) & 1,688 (37.46\%) & \\
         Auto-generated Train &  179,345 & 1,980 & 92,136 (51.37\%) & 87,209 (48.63\%) & \\
         \midrule
         Stage 1 train & 9,012 & 1,915  & \multicolumn{2}{c}{4506 (50\%)} & 4506 (50\%)\\
         \midrule
         Dev &  556 & 52 & 250 (44.96\%) & 213 (38.31\%) & 93 (16.73\%) \\
         Test & 653 & 52 & 274 (41.96\%) & 248 (37.98\%) & 131 (20.06\%) \\
         \bottomrule
    \end{tabular}
    \caption{\semtabfact{} \cite{wang-etal-2021-semeval} statistics. The training data for the first stage was created from the crowdsourced training data using artificial neutral statements created by deleting columns with evidence or swapping statements randomly. For the second stage we use the crowdsourced training data.}
    \label{tab:dataset_stats}
\end{table*}

\section{System}
\label{sec:system}

Our system is a two stage process that first decides whether a statement is neutral, and then decides if non-neutral statements are entailed or refuted.
% Our system is a two stage process that at first decides whether a statement is neutral and in cases it is not, decides whether it is entailed or refuted by the table.
Both stages are implement using a binary \tapas classifier. \tapas \cite{herzig-etal-2020-tapas,eisenschlos-etal-2020-understanding} is a variation of \bert \cite{devlin-etal-2019-bert}, extended with special token embeddings that give the model a notion of the row and column a token is located in and what is its numeric rank with respect to the other cells in the same column.

\subsection{Pre-training}
The original \tapas{} model \cite{herzig-etal-2020-tapas} was pre-trained with a Mask-LM objective \cite{devlin-etal-2019-bert} on tables extracted from Wikipedia.
It was later found \cite{eisenschlos-etal-2020-understanding} that its reasoning capabilities can be improved by further training on artificial counter-factual entailment data.
This led to substantial improvements on the \tabfact{}  dataset \cite{Chen2020TabFact}, a binary table entailment task similar to \semtabfact{}.
On that dataset the test set accuracy for a \bert{}-base-sized model improved from 69.6 to 78.6.
In this work, we use models fine-tuned on \tabfact{} as the foundation for both stages.
We also experimented with using models fine-tuned on \infotabs{} \cite{gupta2020infotabs} and \sqa{} \cite{iyyer-etal-2017-search} as the initial models but did not find that to achieve better accuracy. The overall pre-training strategy is described in Figure~\ref{fig:pipeline}, where we also show how we use these checkpoints to use the two classification models described below.

\subsection{Neutral Identification Stage}

\begin{table*}
\begin{center}
\resizebox{\linewidth}{!}{
\begin{tabular}{ll|cccc|cccc}
\toprule
Stage 1 & Stage 2 & \multicolumn{4}{c}{Dev} & \multicolumn{4}{c}{Test}\\
 &  & \multicolumn{2}{c}{f1 2-way} & \multicolumn{2}{c}{f1 3-way} & \multicolumn{2}{c}{f1 2-way} & \multicolumn{2}{c}{f1 3-way}\\
 &  & Median & Ensemble & Median & Ensemble & Median & Ensemble & Median & Ensemble\\
\midrule
Majority & Majority & \multicolumn{2}{c}{51.44} & \multicolumn{2}{c|}{42.80} & \multicolumn{2}{c}{ 52.41} & \multicolumn{2}{c}{42.15} \\
Majority & \tabfact & \textbf{78.33\err{0.45}} & \textbf{80.25} & 66.40\err{0.66} & 68.29 & \textbf{75.33\err{0.79}} & \textbf{75.21} & 60.64\err{0.65} & 60.47\\
\midrule
\masklm & \tabfact & 74.98\err{0.39} & 78.38 & 70.81\err{0.66} & 72.80 & 74.32\err{0.84} & 74.84 & 67.76\err{0.50} & 67.67\\
\bert & \tabfact & 75.54\err{0.75} & 77.01 & 70.33\err{0.59} & 72.04 & 72.73\err{0.86} & 73.18 & 66.15\err{0.38} & 67.70\\
Inter & \tabfact & 75.77\err{0.50} & 78.28 & \textbf{71.21\err{0.34}} & 72.79 & 72.94\err{0.88} & 74.01 & \textbf{67.99\err{0.78}} & 67.98\\
\midrule
\tabfact (drop) & \tabfact & 78.02\err{0.45} & 80.06 & 67.88\err{0.87} & 69.47 & 74.92\err{0.76} & 75.02 & 62.41\err{0.51} & 61.67\\
\tabfact (random) & \tabfact & 75.97\err{0.73} & 78.50 & 69.62\err{0.93} & 71.81 & 74.77\err{0.97} & 74.67 & 66.64\err{0.16} & 67.11\\
\midrule
\tabfact & \bert & 54.41\err{0.51} & 55.09 & 52.00\err{0.96} & 52.87 & 56.14\err{0.45} & 56.49 & 53.29\err{1.17} & 54.15\\
\tabfact & \masklm & 61.76\err{1.06} & 65.09 & 58.95\err{0.64} & 61.62 & 58.49\err{0.15} & 60.04 & 55.89\err{0.40} & 57.01\\
\tabfact & Inter & 74.00\err{0.32} & 76.68 & 68.86\err{0.48} & 71.33 & 71.08\err{0.78} & 72.14 & 64.94\err{0.34} & 66.43\\
\midrule
\tabfact & \tabfact & 75.74\err{0.18} & 78.33 & 70.76\err{0.55} & \textbf{72.95} & 73.74\err{0.95} & 74.01 & 67.67\err{0.96} & \textbf{68.03}\\
\bottomrule
\end{tabular}
}
\caption{Stage 1 and 2 ablation at 20,000 steps. majority, \tabfact (drop) and \tabfact (random) use majority voting (always predicting non-neutral), only the artificial data created by removing columns and only the random neutral statements respectively. All other models use both kinds of artificial statements. }
\label{tab:res_s1_20k} 
\end{center}
\end{table*}

As discussed, the first stage of the system identifies if a statement is neutral.
Training a system for this task is challenging as the \semtabfact{} training data does not contain neutral statements.
We therefore created artificial neutral statements from two sources.
Following the recommendation of the shared-task organizers, we created neutral statement by randomly pairing statements from the training set with new tables.
Additionally, we created neutral statements by identifying columns that contained evidence for deciding whether a statement is entailed and then randomly removing one of these columns.
Our assumption is that it should not be possible to decide whether the statement is entailed when an evidence column has been removed.
We do not remove the first column of a table since that often contains the name of the row entries.
In order to detect the columns containing the evidence, we trained an ensemble of 5 \tapas QA models on the automatically generated \semtabfact{} training set.
Note that the auto-generated data is generated from templates and in contrast to the crowdsourced training data does have evidence cell annotations.
The models are trained to predict the binary entailment decision as well as the evidence cells at the same time,
and are initialized using a \tapas model fine-tuned on \sqa{}.
The model is trained to predict the binary entailment decision as well as the evidence cells at the same time.
Our models take as input $ [CLS] s_1 ... s_n [SEP] t_1 ... t_{m}$ where $s_1,..., s_n$ represents the tokenized statement and $t_1, ..., t_{m}$ the tokenized table. For each token $t$ of the table the model outputs a score for the token to be an evidence for the statement $S$, $s(t \in S) \in \mathbb{R}$.  Additionally, it outputs the scores of the entailment decision using the $[CLS]$ tokens $s([CLS]) \in \mathbb{R}$.

We use the same hyper-parameters as \sqa{} (as discussed in \citet{herzig-etal-2020-tapas}).
We then run these models over the crowdsourced training data and for all examples where the majority of the models correctly predicts the entailment label, 
we extract all columns for which a majority of the ensemble predicted at least one evidence cell.
Evaluation on the \semtabfact{} development set showed that the precision of this column selection process is 0.87 (87\% of the extracted columns contain a reference cell).
For each column, we then create a new artificial neutral example by removing the respective column from the table.
This procedure yields 651 unique new instances from the 4506 training examples.
However, similarly to the first approach of pairing random statements and tables, the process is not perfect. It may happen that refuted statements continue to be refuted after  removing some of the evidence, but in practice we find it beneficial to generate examples in this fashion.

The final training data is then created by taking the original crowdsourced training examples as positive examples and randomly sampling an equally-sized set of negative examples,
where half of the negatives are random combinations of a statement with a table and the other half are drawn with replacement from the 651 artificial examples.
% Counters from neutral data generation script.
% Statements 4506
% Found row 3575
% Incorrect label: 0 2037
% Columns 1269
% Columns with majority vote 771
% num_row_statements 706
% Columns valid 651
% num_column_statements 651
% Incorrect label: 1 556

\subsection{Entailment Stage}

Training the entailment stage is rather straight-forward, we train the model on the crowdsourced training data
using the same hyper-parameters as \citet{eisenschlos-etal-2020-understanding}.

\subsection{Calibration and Ensemble}

As our training data for stage 1 is balanced but the development data is skewed we find it to improve accuracy if we trigger for examples with a logit larger than 4.0 (rather than 0.0).
Empirically we also find the threshold of 4.0 to work better for the second stage.
This could be explained by the fact that the development set has a different label distribution than the training set.

We train 5 models per stage and use them as an ensemble.
The ensemble score is defined as the median of all the model scores.
Using the median worked better than the mean and voting in preliminary experiments.
\section{Experimental Setup}
\label{sec:experiments}

In this section we explain the \semtabfact task and dataset and give additional details about the experimental setup we used.

The \semtabfact{} dataset consists of statements and tables from the scientific literature.
It is much smaller than similar datasets such as \tabfact{} \cite{Chen2020TabFact} and \infotabs \cite{gupta2020infotabs}.
It is note-worthy that the training set only contains entailed and refuted statements while the dev and test set also contain neutral (unknown) statements.
The statements were written by crowd workers, which presented with 7 different types of statements were instructed to write one statement of each type.
The types of statements were using aggregation, superlatives, counting, comparatives, unique counting and the usage of the caption or common-sense knowledge.

The main metric of the task is the micro f1-score computed over the statements belonging to a table.
The 3-way score takes all statements into account while the 2-way score is restricted to refuted and entailed statements. 

\section{Results}
\label{sec:results}

Table \ref{tab:res_s1_20k} compares our system to multiple baselines.
Unless stated otherwise all baselines have been trained with the same neutral data generation as discussed above and for 20,000 steps.
All numbers are based on 5 independent model runs.
For all setups we report the median of the individual runs as well as the results for a system based on the median logit of the 5 models.
We report error margins for the medians as half the inter-quartile range.

Looking at the first stage of the system in Table \ref{tab:res_s1_20k}, we see that the system based on \tabfact{} is the best choice for the initialization, out-performing a simple \bert{} model as well as models
trained with only the mask-lm and intermediate pre-training on both dev and test ensemble accuracy.
However, the model trained on the intermediate data gives higher median dev and test accuracy (e.g. 72.12 vs 70.76).

With respect to the data generation we observe that any kind of neutral data generation out-performs the majority baseline.
Combining the column removal and random statements yields the best results.
The drop in the 2-way metrics going from the majority Stage 1 model to a learned model is expected as that metric ignores all neutral statements in the eval set.

On the second stage of our system (Table \ref{tab:res_s1_20k}), we see that a \tapas model based on \tabfact outperforms the other baselines by a bigger margin than for Stage 1.
For example, a model based on only \masklm pre-training achieves 57.01 test f1 score while the \tabfact{}-based modle achieves 68.03.
We also found that for this stage there is a more pronounced difference between \bert and \masklm (54.15 vs 57.01) and \masklm and intermediate pre-training (57.01 vs 66.43).

Table \ref{tab:res_steps_th} in the appendix shows the results for different number of steps and thresholds showing that results can be slightly tweaked by tuning them.

\section{Analysis}
\label{sec:analysis}

Table \ref{tab:confusion} shows that the recall and precision on the \emph{neutral} class are 37.6 and 71.4, respectively.
Inspecting some instances of false positives, we find that the system is quite easily fooled; for example classifying the statement ``\emph{The lowest Factor 8 is 0.027}'' as non-neutral for a table that has 5 columns labeled as Factor 1 to 5. False negatives are sometimes caused by failing to map words with typos (``paramters'' vs ``parameters'') or abbreviations (``measurement errors'' vs ``ME'').
Adding harder examples of neutral statements to the training set could potentially further improve the identification.
We also see that the recall on the refuted class (74.3) is lower than the recall of the entailed class (85.2) while there precision values are similar.

% \begin{table}
%     \centering
%     \caption{Confusion for stage 1 on development set. Recall on neutral sentences is }
%     \label{tab:confusion_stage_1}
% \end{table}

\begin{table}
\centering
\resizebox{1.0\columnwidth}{!}{
\begin{tabular}{l|rrr}
\toprule
\diagbox[]{Reference}{Prediction} &  Non-neutral & Neutral & Recall \\
\midrule
Non-neutral & 449 & 14 & 97.0 \\
Neutral & 58 & 35 & 37.6 \\
Precision & 88.6 & 71.4\\
\toprule
\diagbox[]{Reference}{Prediction} &    Refuted   &  Entailed  & Recall \\
\midrule
Refuted & 153 & 53 & 74.3 \\
Entailed & 36 & 207 & 85.2 \\
Precision & 81.0 & 79.6 \\
\bottomrule
\end{tabular}
}
\caption{Confusion matrix for Stage 1 and Stage 2 on the development set.}
\label{tab:confusion}
\vspace{-2ex}
\end{table}

In Table~\ref{tab:slices} we construct mutually excluded groups of the validation set. 
Each set is identified by specific keywords appearing in the statement, for example \emph{Comparatives} must contain ``higher'', ``better'', ``than'', etc.
The full list is defined in the appendix of \citet{eisenschlos-etal-2020-understanding}. 
We observe that comparatives and aggregations have the largest total error rates, meaning that the biggest gains in overall accuracy can be made by improving those reasoning skills.
Between these two, Comparatives have the lowest in-group accuracy.
Table \ref{tab:slices1} and Table \ref{tab:slices2} in the appendix show the some anaylsis for Stage 1 and Stage 2, respectively.
The trend for Stage 2 is similar to the overall trend whereas Stage 1 accuracy is relatively stable across the different groups except for comparatives where the accuracy drops from 87\% overall to 81\%. 

Another class of examples with relatively low accuracy are statements around unique counting. We find that statements containing the word \emph{different} have an accuracy of 51.3 (vs. 71\% overall) and account for 3.4 percentage points of the total error rate. Examples include ``\emph{There are six different classes}'' and ``\emph{They have ten different parameters}''.

\begin{table}[!t]
\small
\centering
% \resizebox{1.0\columnwidth}{!}{
\begin{tabular}{lr|rrr} \toprule			
	& Size	&	Acc	&	Baseline	& ER\\
\midrule
Overall               & $100.0$ & $71.0$ & $45.0$ & $29.0$ \\
\midrule
Superlatives          & $15.8$ & $73.9$ & $50.0$ & $4.1$ \\
Aggregations          & $13.8$ & $61.0$ & $46.8$ & $5.4$ \\
Comparatives          & $12.2$ & $58.8$ & $47.1$ & $5.0$ \\
Negations             & $3.1$ & $82.4$ & $41.2$ & $0.5$ \\
\midrule
Multiple of the above & $5.9$ & $72.7$ & $63.6$ & $1.6$ \\
Other                 & $49.1$ & $75.1$ & $43.6$ & $12.2$ \\
\bottomrule
\end{tabular}
% }
\caption{Accuracy and total error rate (ER) for different question groups derived from the same word heuristics defined in ~\citet{eisenschlos-etal-2020-understanding}. The baseline is simple class majority and 
the error rate in each group is taken with respect to the full set.
Comparatives show the biggest margin for future improvements comparing with the overall system accuracy.
}
\vspace{-4ex}
\label{tab:slices}
\end{table}

\section{Conclusion}
\label{sec:conclusion}

We presented our contribution to the \semtabfact{} task \cite{wang-etal-2021-semeval} on table entailment. Our system consists of two stages that classify statements into non-neutral or neutral and refuted or entailed. 
Our model achieves 68.03 average micro f1-score on the test set. 
We showed that our procedure for creating artificial neutral statements improves the system over a majority baseline but results in a relatively low recall of 37.6. Other methods for creating harder neutral statements might further improve this value.
In line with \citet{eisenschlos-etal-2020-understanding}, we find that pre-training on intermediate data improves the system accuracy over a system purely pre-trained with a \masklm{} objective.
While these initial results look promising, we find that the model struggles with statements that involve complex operations such as comparisons and unique counting.

\bibliographystyle{style/acl_natbib}
\bibliography{bib/journal-full,bib/jbg}

\begin{thebibliography}{29}
\expandafter\ifx\csname natexlab\endcsname\relax\def\natexlab#1{#1}\fi

\bibitem[{Alberti et~al.(2019)Alberti, Andor, Pitler, Devlin, and
  Collins}]{alberti-etal-2019-synthetic}
Chris Alberti, Daniel Andor, Emily Pitler, Jacob Devlin, and Michael Collins.
  2019.
\newblock \href {https://doi.org/10.18653/v1/P19-1620} {Synthetic {QA} corpora
  generation with roundtrip consistency}.
\newblock In \emph{Proceedings of the Association for Computational
  Linguistics}, pages 6168--6173, Florence, Italy. Association for
  Computational Linguistics.

\bibitem[{Bowman et~al.(2015)Bowman, Angeli, Potts, and
  Manning}]{bowman-etal-2015-large}
Samuel~R. Bowman, Gabor Angeli, Christopher Potts, and Christopher~D. Manning.
  2015.
\newblock \href {https://doi.org/10.18653/v1/D15-1075} {A large annotated
  corpus for learning natural language inference}.
\newblock In \emph{Proceedings of Empirical Methods in Natural Language
  Processing}, pages 632--642, Lisbon, Portugal. Association for Computational
  Linguistics.

\bibitem[{Chen et~al.(2020)Chen, Wang, Chen, Zhang, Wang, Li, Zhou, and
  Wang}]{Chen2020TabFact}
Wenhu Chen, Hongmin Wang, Jianshu Chen, Yunkai Zhang, Hong Wang, Shiyang Li,
  Xiyou Zhou, and William~Yang Wang. 2020.
\newblock \href {https://openreview.net/forum?id=rkeJRhNYDH} {Tabfact: A
  large-scale dataset for table-based fact verification}.
\newblock In \emph{Proceedings of the International Conference on Learning
  Representations}.

\bibitem[{Dagan et~al.(2010)Dagan, Dolan, Magnini, and Roth}]{dagan-10}
Ido Dagan, Bill Dolan, Bernardo Magnini, and Dan Roth. 2010.
\newblock Recognizing textual entailment: Rationale, evaluation and approaches.
\newblock \emph{Journal of Natural Language Engineering}, 4.

\bibitem[{Dagan et~al.(2005)Dagan, Glickman, and Magnini}]{dagan2005pascal}
Ido Dagan, Oren Glickman, and Bernardo Magnini. 2005.
\newblock The pascal recognising textual entailment challenge.
\newblock In \emph{Machine Learning Challenges Workshop}, pages 177--190.
  Springer.

\bibitem[{Devlin et~al.(2019)Devlin, Chang, Lee, and
  Toutanova}]{devlin-etal-2019-bert}
Jacob Devlin, Ming-Wei Chang, Kenton Lee, and Kristina Toutanova. 2019.
\newblock \href {https://doi.org/10.18653/v1/N19-1423} {{BERT}: Pre-training of
  deep bidirectional transformers for language understanding}.
\newblock In \emph{Conference of the North American Chapter of the Association
  for Computational Linguistics}, pages 4171--4186, Minneapolis, Minnesota.
  Association for Computational Linguistics.

\bibitem[{Eisenschlos et~al.(2020)Eisenschlos, Krichene, and
  M{\"u}ller}]{eisenschlos-etal-2020-understanding}
Julian Eisenschlos, Syrine Krichene, and Thomas M{\"u}ller. 2020.
\newblock \href {https://www.aclweb.org/anthology/2020.findings-emnlp.27}
  {Understanding tables with intermediate pre-training}.
\newblock In \emph{Findings of the Association for Computational Linguistics:
  EMNLP}, pages 281--296, Online. Association for Computational Linguistics.

\bibitem[{Gardner et~al.(2020)Gardner, Artzi, Basmov, Berant, Bogin, Chen,
  Dasigi, Dua, Elazar, Gottumukkala, Gupta, Hajishirzi, Ilharco, Khashabi, Lin,
  Liu, Liu, Mulcaire, Ning, Singh, Smith, Subramanian, Tsarfaty, Wallace,
  Zhang, and Zhou}]{gardner-2020}
Matt Gardner, Yoav Artzi, Victoria Basmov, Jonathan Berant, Ben Bogin, Sihao
  Chen, Pradeep Dasigi, Dheeru Dua, Yanai Elazar, Ananth Gottumukkala, Nitish
  Gupta, Hannaneh Hajishirzi, Gabriel Ilharco, Daniel Khashabi, Kevin Lin,
  Jiangming Liu, Nelson~F. Liu, Phoebe Mulcaire, Qiang Ning, Sameer Singh,
  Noah~A. Smith, Sanjay Subramanian, Reut Tsarfaty, Eric Wallace, Ally Zhang,
  and Ben Zhou. 2020.
\newblock \href {https://doi.org/10.18653/v1/2020.findings-emnlp.117}
  {Evaluating models{'} local decision boundaries via contrast sets}.
\newblock In \emph{Findings of the Association for Computational Linguistics:
  EMNLP}, pages 1307--1323, Online. Association for Computational Linguistics.

\bibitem[{Geva et~al.(2020)Geva, Gupta, and Berant}]{Geva2020InjectingNR}
Mor Geva, Ankit Gupta, and Jonathan Berant. 2020.
\newblock \href {https://doi.org/10.18653/v1/2020.acl-main.89} {Injecting
  numerical reasoning skills into language models}.
\newblock In \emph{Proceedings of the Association for Computational
  Linguistics}, pages 946--958, Online. Association for Computational
  Linguistics.

\bibitem[{Gupta et~al.(2020)Gupta, Mehta, Nokhiz, and
  Srikumar}]{gupta2020infotabs}
Vivek Gupta, Maitrey Mehta, Pegah Nokhiz, and Vivek Srikumar. 2020.
\newblock \href {https://arxiv.org/abs/2005.06117} {Infotabs: Inference on
  tables as semi-structured data}.
\newblock In \emph{Proceedings of the Association for Computational
  Linguistics}, Seattle, Washington. Association for Computational Linguistics.

\bibitem[{Gururangan et~al.(2020)Gururangan, Marasovi{\'c}, Swayamdipta, Lo,
  Beltagy, Downey, and Smith}]{gururangan-2020-dont-stop}
Suchin Gururangan, Ana Marasovi{\'c}, Swabha Swayamdipta, Kyle Lo, Iz~Beltagy,
  Doug Downey, and Noah~A. Smith. 2020.
\newblock \href {https://arxiv.org/abs/2004.10964} {Don't stop pretraining:
  Adapt language models to domains and tasks}.
\newblock In \emph{Proceedings of the Association for Computational
  Linguistics}, Seattle, Washington. Association for Computational Linguistics.

\bibitem[{Herzig et~al.(2020)Herzig, Nowak, M{\"u}ller, Piccinno, and
  Eisenschlos}]{herzig-etal-2020-tapas}
Jonathan Herzig, Pawel~Krzysztof Nowak, Thomas M{\"u}ller, Francesco Piccinno,
  and Julian Eisenschlos. 2020.
\newblock \href {https://doi.org/10.18653/v1/2020.acl-main.398} {{T}a{P}as:
  Weakly supervised table parsing via pre-training}.
\newblock In \emph{Proceedings of the Association for Computational
  Linguistics}, pages 4320--4333, Online. Association for Computational
  Linguistics.

\bibitem[{Howard and Ruder(2018)}]{howard-2018-universal}
Jeremy Howard and Sebastian Ruder. 2018.
\newblock \href {https://doi.org/10.18653/v1/P18-1031} {Universal language
  model fine-tuning for text classification}.
\newblock In \emph{Proceedings of the Association for Computational
  Linguistics}, pages 328--339, Melbourne, Australia. Association for
  Computational Linguistics.

\bibitem[{Iyer et~al.(2017)Iyer, Dandekar, , and Csernai}]{iyer-17}
Shankar Iyer, Nikhil Dandekar, , and Korn\'el Csernai. 2017.
\newblock \href
  {https://www.quora.com/q/quoradata/First-Quora-Dataset-Release-Question-Pairs}
  {Quora question pairs}.

\bibitem[{Iyyer et~al.(2017)Iyyer, Yih, and Chang}]{iyyer-etal-2017-search}
Mohit Iyyer, Wen-tau Yih, and Ming-Wei Chang. 2017.
\newblock \href {https://doi.org/10.18653/v1/P17-1167} {Search-based neural
  structured learning for sequential question answering}.
\newblock In \emph{Proceedings of the Association for Computational
  Linguistics}, pages 1821--1831, Vancouver, Canada. Association for
  Computational Linguistics.

\bibitem[{Jo et~al.(2019)Jo, Trummer, Yu, Wang, Yu, Liu, and
  Mehta}]{jo2019aggchecker}
Saehan Jo, Immanuel Trummer, Weicheng Yu, Xuezhi Wang, Cong Yu, Daniel Liu, and
  Niyati Mehta. 2019.
\newblock \href {https://doi.org/10.14778/3352063.3352104} {Aggchecker: A
  fact-checking system for text summaries of relational data sets}.
\newblock \emph{International Conference on Very Large Databases},
  12(12):1938–1941.

\bibitem[{Kaushik et~al.(2020)Kaushik, Hovy, and Lipton}]{Kaushik2020Learning}
Divyansh Kaushik, Eduard~H. Hovy, and Zachary~Chase Lipton. 2020.
\newblock \href {https://openreview.net/forum?id=Sklgs0NFvr} {Learning the
  difference that makes {A} difference with counterfactually-augmented data}.
\newblock In \emph{Proceedings of the International Conference on Learning
  Representations}, Addis Ababa, Ethiopia.

\bibitem[{Leonandya et~al.(2019)Leonandya, Hupkes, Bruni, and
  Kruszewski}]{leonandya-etal-2019-fast}
Rezka Leonandya, Dieuwke Hupkes, Elia Bruni, and Germ{\'a}n Kruszewski. 2019.
\newblock \href {https://doi.org/10.18653/v1/W19-0419} {The fast and the
  flexible: Training neural networks to learn to follow instructions from small
  data}.
\newblock In \emph{Proceedings of the International Conference on Computational
  Semantics}, pages 223--234, Gothenburg, Sweden. Association for Computational
  Linguistics.

\bibitem[{Lewis et~al.(2019)Lewis, Denoyer, and
  Riedel}]{lewis-etal-2019-unsupervised}
Patrick Lewis, Ludovic Denoyer, and Sebastian Riedel. 2019.
\newblock \href {https://doi.org/10.18653/v1/P19-1484} {Unsupervised question
  answering by cloze translation}.
\newblock In \emph{Proceedings of the Association for Computational
  Linguistics}, pages 4896--4910, Florence, Italy. Association for
  Computational Linguistics.

\bibitem[{Pruksachatkun et~al.(2020)Pruksachatkun, Phang, Liu, Htut, Zhang,
  Pang, Vania, Kann, and Bowman}]{pruksachatkun-2020-intermediate-task}
Yada Pruksachatkun, Jason Phang, Haokun Liu, Phu~Mon Htut, Xiaoyi Zhang,
  Richard~Yuanzhe Pang, Clara Vania, Katharina Kann, and Samuel~R. Bowman.
  2020.
\newblock \href {https://arxiv.org/abs/2005.00628} {Intermediate-task transfer
  learning with pretrained models for natural language understanding: When and
  why does it work?}
\newblock In \emph{Proceedings of the Association for Computational
  Linguistics}, Seattle, Washington. Association for Computational Linguistics.

\bibitem[{Salvatore et~al.(2019)Salvatore, Finger, and
  Hirata~Jr}]{salvatore-etal-2019-logical}
Felipe Salvatore, Marcelo Finger, and Roberto Hirata~Jr. 2019.
\newblock \href {https://doi.org/10.18653/v1/D19-6103} {A logical-based corpus
  for cross-lingual evaluation}.
\newblock In \emph{Proceedings of the 2nd Workshop on Deep Learning Approaches
  for Low-Resource NLP (DeepLo 2019)}, pages 22--30, Hong Kong, China.
  Association for Computational Linguistics.

\bibitem[{Suhr et~al.(2017)Suhr, Lewis, Yeh, and Artzi}]{suhr2017corpus}
Alane Suhr, Mike Lewis, James Yeh, and Yoav Artzi. 2017.
\newblock \href {https://doi.org/10.18653/v1/P17-2034} {A corpus of natural
  language for visual reasoning}.
\newblock In \emph{Proceedings of the Association for Computational
  Linguistics}, pages 217--223, Vancouver, Canada. Association for
  Computational Linguistics.

\bibitem[{Suhr et~al.(2019)Suhr, Zhou, Zhang, Zhang, Bai, and
  Artzi}]{suhr2019corpus}
Alane Suhr, Stephanie Zhou, Ally Zhang, Iris Zhang, Huajun Bai, and Yoav Artzi.
  2019.
\newblock \href {https://doi.org/10.18653/v1/P19-1644} {A corpus for reasoning
  about natural language grounded in photographs}.
\newblock In \emph{Proceedings of the Association for Computational
  Linguistics}, pages 6418--6428, Florence, Italy. Association for
  Computational Linguistics.

\bibitem[{Vlachos and Riedel(2015)}]{vlachos-riedel-2015-identification}
Andreas Vlachos and Sebastian Riedel. 2015.
\newblock \href {https://doi.org/10.18653/v1/D15-1312} {Identification and
  verification of simple claims about statistical properties}.
\newblock In \emph{Proceedings of Empirical Methods in Natural Language
  Processing}, pages 2596--2601, Lisbon, Portugal. Association for
  Computational Linguistics.

\bibitem[{Wang et~al.(2021)Wang, Mahajan, Danilevsky, and
  Rosenthal}]{wang-etal-2021-semeval}
Nancy Xin~Ru Wang, Diwakar Mahajan, Marina Danilevsky, and Sara Rosenthal.
  2021.
\newblock Semeval-2021 task 9: Fact verification and evidence finding for
  tabular data in scientific documents (sem-tab-facts).
\newblock In \emph{Proceedings of the 15th international workshop on semantic
  evaluation (SemEval-2021)}.

\bibitem[{Wang et~al.(2015)Wang, Berant, and Liang}]{wang-etal-2015-building}
Yushi Wang, Jonathan Berant, and Percy Liang. 2015.
\newblock \href {https://doi.org/10.3115/v1/P15-1129} {Building a semantic
  parser overnight}.
\newblock In \emph{Proceedings of the Association for Computational
  Linguistics}, pages 1332--1342, Beijing, China. Association for Computational
  Linguistics.

\bibitem[{Weir et~al.(2020)Weir, Utama, Galakatos, Crotty, Ilkhechi, Ramaswamy,
  Bhushan, Geisler, H\"{a}ttasch, Eger, Cetintemel, and Binnig}]{dbpal}
Nathaniel Weir, Prasetya Utama, Alex Galakatos, Andrew Crotty, Amir Ilkhechi,
  Shekar Ramaswamy, Rohin Bhushan, Nadja Geisler, Benjamin H\"{a}ttasch,
  Steffen Eger, Ugur Cetintemel, and Carsten Binnig. 2020.
\newblock \href {https://doi.org/10.1145/3318464.3380589} {Dbpal: A fully
  pluggable nl2sql training pipeline}.
\newblock In \emph{Proceedings of the ACM SIGMOD International Conference on
  Management of Data}, SIGMOD ’20, page 2347–2361, New York, NY, USA.
  Association for Computing Machinery.

\bibitem[{Wu et~al.(2016)Wu, Shi, Chen, Huang, and
  Su}]{wu-etal-2016-bilingually}
Changxing Wu, Xiaodong Shi, Yidong Chen, Yanzhou Huang, and Jinsong Su. 2016.
\newblock \href {https://doi.org/10.18653/v1/D16-1253} {Bilingually-constrained
  synthetic data for implicit discourse relation recognition}.
\newblock In \emph{Proceedings of Empirical Methods in Natural Language
  Processing}, pages 2306--2312, Austin, Texas. Association for Computational
  Linguistics.

\bibitem[{Zhang et~al.(2020)Zhang, Wang, Wang, Cao, Zhang, and
  Wang}]{zhang-etal-2020-table}
Hongzhi Zhang, Yingyao Wang, Sirui Wang, Xuezhi Cao, Fuzheng Zhang, and
  Zhongyuan Wang. 2020.
\newblock \href {https://www.aclweb.org/anthology/2020.emnlp-main.126} {Table
  fact verification with structure-aware transformer}.
\newblock In \emph{Proceedings of Empirical Methods in Natural Language
  Processing}, pages 1624--1629, Online. Association for Computational
  Linguistics.

\end{thebibliography}

\clearpage

\appendix\section*{Appendix}

The appendix contains additional results and analysis tables.

\section{Results}

Table \ref{tab:res_steps_th} shows the results for different number of steps and thresholds showing that results can be slightly tweaked by tuning them.

\begin{table}[h]
\begin{center}
\resizebox{1.0\linewidth}{!}{
\begin{tabular}{ll|cccc}
\toprule
Steps & Thresh & \multicolumn{2}{c}{f1 2-way} & \multicolumn{2}{c}{f1 3-way}\\
 &  & Median & Ensemble & Median & Ensemble\\
\midrule
20K & 0.0 & 74.97\err{0.35} & 76.41 & 69.96\err{1.05} & 71.03\\
10K & 4.0 & \textbf{75.97\err{1.48}} & 76.47 & 70.68\err{1.36} & 72.19\\
10K & 0.0 & 75.84\err{1.33} & 76.55 & 70.63\err{1.21} & 72.27\\
20K & 4.0 & 75.74\err{0.18} & \textbf{78.33} & \textbf{70.76\err{0.55}} & \textbf{72.95}\\
\bottomrule
\end{tabular}
}
\caption{Ablation of steps and threshold on the dev set.}
\label{tab:res_steps_th} 
\end{center}
\end{table}

\section{Analysis}

Table \ref{tab:slices1} and Table \ref{tab:slices2} show the error rate contributions of different types of statements for Stage 1 and Stage 2, respectively.
The trend for Stage 2 is similar to the overall trend (Table \ref{tab:slices}) whereas Stage 1 accuracy is relatively stable across the different groups except for comparatives where the accuracy drops from 87\% overall to 81\%. 

\begin{table}[h]
\small
\centering
% \resizebox{1.0\columnwidth}{!}{
\begin{tabular}{lr|rrr} \toprule			
	& \textbf{Size}	&	\textbf{Acc}	&	\textbf{Baseline}	& \textbf{ER}\\
\midrule
\textbf{Overall}               & $100.0$ & $87.1$ & $83.3$ & $12.9$ \\
\midrule
\textbf{Superlatives}          & $15.8$ & $90.9$ & $89.8$ & $1.4$ \\
\textbf{Aggregations}          & $13.8$ & $88.3$ & $87.0$ & $1.6$ \\
\textbf{Comparatives}          & $12.2$ & $80.9$ & $79.4$ & $2.3$ \\
\textbf{Negations}             & $3.1$ & $88.2$ & $64.7$ & $0.4$ \\
\midrule
\textbf{Multiple of the above} & $5.9$ & $93.9$ & $87.9$ & $0.4$ \\
\textbf{Other}                 & $49.1$ & $86.1$ & $81.7$ & $6.8$ \\
\bottomrule
\end{tabular}
% }
\caption{Accuracy and total error rate (ER) for different question groups for Stage 1.}
\label{tab:slices1}
\end{table}

\begin{table}[h]
\small
\centering
% \resizebox{1.0\columnwidth}{!}{
\begin{tabular}{lr|rrr} \toprule			
	& \textbf{Size}	&	\textbf{Acc}	&	\textbf{Baseline}	& \textbf{ER}\\
\midrule
\textbf{Overall}               & $100.0$ & $80.2$ & $54.1$ & $19.8$ \\
\midrule
\textbf{Superlatives}          & $16.9$ & $80.3$ & $53.9$ & $3.3$ \\
\textbf{Aggregations}          & $14.0$ & $66.7$ & $54.0$ & $4.7$ \\
\textbf{Comparatives}          & $11.6$ & $71.2$ & $59.6$ & $3.3$ \\
\textbf{Negations}             & $2.4$ & $90.9$ & $63.6$ & $0.2$ \\
\midrule
\textbf{Multiple of the above} & $6.2$ & $75.0$ & $71.4$ & $1.6$ \\
\textbf{Other}                 & $48.8$ & $86.3$ & $53.0$ & $6.7$ \\
\bottomrule
\end{tabular}
% }
\caption{Accuracy and total error rate (ER) for different question groups for Stage 2.}
%\vspace{-3ex}
\label{tab:slices2}
\end{table}

\end{document}